\pdfoutput=1
\documentclass[conference,compsoc]{IEEEtran}
\usepackage{amsmath,epsfig,bm,multirow,caption}
\usepackage[numbers,sort&compress]{natbib}

\begin{document}

\title{E$^2$BoWs: An End-to-End Bag-of-Words Model via Deep Convolutional Neural Network}



\author{\IEEEauthorblockN{Xiaobin Liu\IEEEauthorrefmark{1},
Shiliang Zhang\IEEEauthorrefmark{1},
Tiejun Huang\IEEEauthorrefmark{1},
Qi Tian\IEEEauthorrefmark{2}
\IEEEauthorblockA{\IEEEauthorrefmark{1}School of Electronics Engineering and Computer Science, Peking University, Beijing, 100871, China\\
Email: \{xbliu.vmc, slzhang.jdl, tjhuang\}@pku.edu.cn}
\IEEEauthorblockA{\IEEEauthorrefmark{2}Department of Computer Science, University of Texas at San Antonio, San Antonio, TX 78249-1604, USA\\
Email: qitian@cs.utsa.edu}
}}

\maketitle

\begin{abstract}
Traditional Bag-of-visual Words (BoWs) model is commonly generated with many steps including local feature extraction, codebook generation, and feature quantization, \emph{etc.} Those steps are relatively independent with each other and are hard to be jointly optimized. Moreover, the dependency on hand-crafted local feature makes BoWs model not effective in conveying high-level semantics. These issues largely hinder the performance of BoWs model in large-scale image applications. To conquer these issues, we propose an End-to-End BoWs (E$^2$BoWs) model based on Deep Convolutional Neural Network (DCNN). Our model takes an image as input, then identifies and separates the semantic objects in it, and finally outputs the visual words with high semantic discriminative power. Specifically, our model firstly generates Semantic Feature Maps (SFMs) corresponding to different object categories through convolutional layers, then introduces Bag-of-Words Layers (BoWL) to generate visual words for each individual feature map. We also introduce a novel learning algorithm to reinforce the sparsity of the generated E$^2$BoWs model, which further ensures the time and memory efficiency. We evaluate the proposed E$^2$BoWs model on several image search datasets including \emph{CIFAR-10}, \emph{CIFAR-100}, \emph{MIRFLICKR-25K} and \emph{NUS-WIDE}. Experimental results show that our method achieves promising accuracy and efficiency compared with recent deep learning based retrieval works.
\end{abstract}


\IEEEpeerreviewmaketitle

\section{Introduction}
\label{sec:intro}

\begin{figure*}[t]
\begin{minipage}[b]{1.0\linewidth}
  \centering
  \centerline{\epsfig{figure=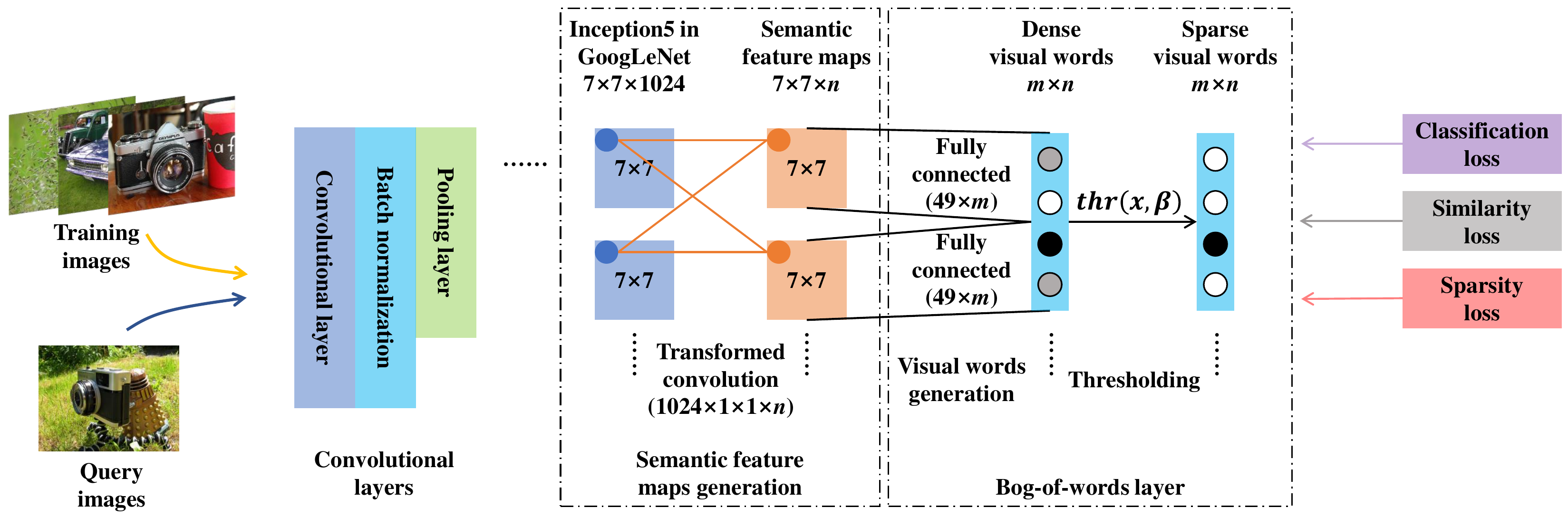,width=14cm}}
\end{minipage}
\captionsetup{belowskip=-10pt}
\caption{Framework of the proposed E$^2$BoWs model. The structure of our deep model is identical to the one of GoogLeNet~\cite{googlenet} with BN~\cite{bn} till the Inception5 layer. The output size of Inception5 is $7 \times 7 \times 1024$. Pool5 in GoogLeNet~\cite{googlenet} is discarded. The $n$-way output layer is transformed into a convolutional layer to generate $n$ semantic feature maps. $m$ sparse visual words are then generated by bog-of-words layer from each individual semantic feature map, resulting in $m \times n$ visual words. Finally, a three-component loss function is applied for training the model.}
\label{fig:structure}
\end{figure*}

A huge number of images are being uploaded to the Internet every moment, and each image commonly conveys rich information. This makes Content-Based Image Retrieval (CBIR) a challenging and promising task. Bag-of-visual Words (BoWs) model, which considers an image as a collection of visual words, has been widely applied for large-scale image retrieval. Conventional BoWs model is computed with many stages, \emph{e.g.}, feature extraction, codebook generation, and feature quantization~\cite{vocabularytree,bundling,videogoogle,discovering}. Then inverted file index and Term Frequency-Inverse Document Frequency (TF-IDF) strategy can be used for indexing and retrieval. Since the number of visual vocabulary is commonly very large, \emph{e.g.}, 1 million in \cite{vocabularytree}, and a certain image only contains a small number of visual words, indexes generated by BoWs model are sparse and thus ensure the high retrieval efficiency.

Most of existing BoWs models are based on hand-crafted local features, \emph{e.g.}, SIFT~\cite{sift}. These models have shown promising performance in large-scale partial-duplicate image retrieval~\cite{vocabularytree,bundling,videogoogle}. However, as the local descriptor cannot effectively describe high-level semantics, \emph{i.e.}, commonly known as the ``semantic gap" issue, BoWs models build on local descriptors always fail to address the semantic similar image retrieval task~\cite{wan}. Although some works have been proposed to conquer this issue~\cite{semantics,enhancing,semanticaware}, most of these works introduce extra computations and memory overheads.

Recent years have witnessed a lot of breakthroughs in end-to-end deep learning model for vision tasks. After AlexNet~\cite{alexnet} achieving the best performance in ImageNet Large-Scale Visual Recognition Challenge (ILSVRC), Deep Convolutional Neural Network (DCNN) has been applied to various vision tasks, including image classification~\cite{googlenet, resnet}, object detection~\cite{rcnn, fastrcnn}, semantic segmentation~\cite{fullyconv} and many other tasks~\cite{facenet2015,wang2014learning,places2,deepid2}. Most of DCNNs consist of a set of convolutional layers and Fully Connected (FC) layers. It is found that convolutional layers can extract high-level semantic cues from pixel-level input and hence provide a possible solution to solve the ``semantic ga'' issue. Therefore, it is straightforward to leverage DCNN in image retrieval~\cite{wan}. Some works use DCNN to generate hash codes and yield promising performance~\cite{dsh, cnnh, linkevin,zhao2015}. However, there still lacks research efforts on DCNN based BoWs model, which could be integrated with inverted file indexing and TF-IDF weighting for large-scale image retrieval.

Targeting to leverage the efficiency of BoWs model and the semantic learning ability of DCNN models in large-scale image retrieval, we propose to generate a DCNN based End-to-End BoWs (E$^2$BoWs) model as shown in Fig. \ref{fig:structure}. Structure of our E$^2$BoWs model coincides with GoogLeNet~\cite{googlenet} with Batch Normalization (BN)~\cite{bn} up to Inception5. We discard Pool5 layer and transform the last FC layer into a convolutional layer to generate Semantic Feature Maps (SFMs) specifically corresponding to different object categories. A Bag-of-Words Layer (BoWL) is then introduced to generate sparse visual words from each semantic feature map. This ensures the resulting visual words to preserve clear semantic cues. Finally, a three-component loss function is designed to ensure: 1) fast convergence of the training procedure, 2) similar images sharing more visual words, and 3) high sparsity of the generated E$^2$BoWs model, respectively.

The proposed method has several advantages compared with traditional BoWs models: 1) Instead of using hand-crafted features and being generated with several steps, our E$^2$BoWs model is generated in an end-to-end manner, thus is more efficient and easier to be jointly optimized and tuned. 2) Incorporating DCNN into BoWs model is potential to bring higher discriminative power to semantics and provide a better solution for semantic similar image search task. Our E$^2$BoWs model also shows advantages over traditional hashing methods in that it conveys clear semantic cues. We evaluate the proposed E$^2$BoWs model on several image search datasets including \emph{CIFAR-10}, \emph{CIFAR-100}, \emph{MIRFLICKR-25K}, and \emph{NUS-WIDE}. Comparisons with recent deep learning based image retrieval works show that our method achieves promising accuracy and efficiency.

The rest of this paper is organized as follows: Section \ref{sec:related} discusses some works related to our model. Section \ref{sec:model} presents our model in detail. Section \ref{sec:exper} evaluates the proposed model on different datasets and Section \ref{sec:concl} gives our conclusions.

\section{Related Work}
\label{sec:related}

As a fundamental task in multimedia content analysis and computer vision~\cite{content,wan,visualrank}, CBIR aims to search for images similar with the query in an image gallery. Since directly computing similarity between two images with raw image pixels is infeasible, BoWs model is widely used as an image representation for large-scale image retrieval. Over the past decade, various BoWs models~\cite{vocabularytree,bundling,videogoogle,discovering} have been proposed based on local descriptors, such as SIFT~\cite{sift} and SURF~\cite{surf}. Those BoWs models have shown promising performance in large-scale image retrieval. Conventional BoWs models consider an image as a collection of visual words and is generated by many stages, \emph{e.g.}, feature extraction, codebook generation and feature quantization~\cite{vocabularytree,bundling,videogoogle,discovering}. For instnace, Nister~\emph{et al.}~\cite{vocabularytree} extract SIFT~\cite{sift} descriptors from MSER regions~\cite{mser} and then hierarchically quantize SIFT descriptors by the vocabulary tree. As individual visual word cannot depict the spatial cues in images, some works combine visual words with spatial information~\cite{spatial,integrated} to make the resulting BoWs model more discriminative to the spatial cues. Some other works aim to generate more effective and discriminative vocabularies~\cite{codebook,universal}.

However, the dependency on hand-crafted local feature hinders the ability of conventional visual words to convey semantic cues due to the ``semantic gap" between low-level local features and high-level semantics. For instance, two objects from different categories might share similar local features, which can be quantized to same visual words in the vocabulary tree.

Some works have been proposed to enhance the discriminative power of BoWs model to semantic cues~\cite{semantics,enhancing,semanticaware}. Wu \emph{et al.}~\cite{semantics} propose an off-line distance metric learning scheme to map related features to the same visual words to generate an optimized codebook. Wu \emph{et al.}~\cite{enhancing} present an on-line metric learning algorithm to improve the BoWs model by optimizing the proposed semantic loss. Zhang \emph{et al.}~\cite{semanticaware} propose a method to co-index semantic attributes into inverted index generated by local features to make it convey more semantic cues. However, most of these works need extra computations either in the off-line indexing or on-line retrieval stages. Moreover, since these models are generated by many independent steps, they are hard to be jointly optimized to achieve better efficiency and accuracy.

Recently, many works leverage DCNN in CBIR~\cite{wan,dsh, cnnh, linkevin,zhao2015,lai,zhu}. Wan \emph{et al.}~\cite{wan} propose three schemes to apply DCNN in CBIR, \emph{i.e.}, 1) directly use the features from the model pre-trained on \emph{ImageNet}~\cite{imagenet}, 2) refine the features by metric learning, and 3) retrain the model on the domain dataset. They prove that DCNN based features can significantly outperform hand-crafted features after being fine-tuned. However, they don't consider the retrieval efficiency when apply the features in large-scale datasets. Xia \emph{et al.}~\cite{cnnh} introduce a DCNN based hashing method. The method consists of two steps: first generate hash codes on training set by an iterative algorithm, and then learn a hash function based on DCNN to fit the hash codes generated in step 1. The independence of two steps hinders the joint learning of the whole model. Lin \emph{et al.}~\cite{linkevin} propose a framework to generate hash codes directly by a classification object function. They show that deep model trained by classification task can be adopted for CBIR task. Zhao \emph{et al.}~\cite{zhao2015} and Lai \emph{et al.}~\cite{lai} use triplet loss to train the network to preserve semantic relations of images.

In these aforementioned methods, real-value hash codes are learned during training. The real-value hash codes are then quantized to binary codes for testing. Different distance metrics used in training and testing,\emph{ e.g.}, Euclidean distance and Hamming distance, may bring approximation error and hinder the training efficiency. Quantization error could also be produced by the quantization stage. Different from those works, Liu \emph{et al.}~\cite{dsh} and Zhu \emph{et al.}~\cite{zhu} reinforce the networks to output binary-like hash codes to reduce quantization error and approximation error. So far, most of deep learning based retrieval works focus on generating hashing codes. There still lacks research efforts in DCNN based BoWs model. It is promising to generate a discriminative BoWs model directly from an end-to-end DCNN and leverage the scalability of BoWs model for large-scale image retrieval.

\section{Proposed Method}
\label{sec:model}

E$^2$BoWs model is generated by modifying the GoogLeNet~\cite{googlenet} with BN~\cite{bn}. As shown in Fig. \ref{fig:structure}, before the Inception5 layer, the structure of our deep model is identical to the one of GoogLeNet~\cite{googlenet} with BN~\cite{bn}. Most of previous works extract features for retrieval from FC layers. Differently, we propose to learn features from feature maps which preserve more visual cues than FC layers. We thus transform the last $n$-way FC layer into a convolutional layer to generate $n$ SFMs corresponding to $n$ training categories. Then, $m$ sparse visual words are generated from each individual SFM by the Bag-of-Words Layer, resulting in $m \times n$ visual words. Finally, a three-component loss function is applied to train the model. In the following parts, we present the details of the network structure, model training and generalization ability improvement.

\subsection{Semantic Feature Maps Generation}

In GoogLeNet~\cite{googlenet}, the output layer conveys semantic cues because the label supervision is directly applied on it. However, the output layer losses certain visual details of the images, such as the location and size of objects, which could be beneficial in image retrieval. Meantime, Inception5 contains more visual cues than semantics. Learning visual words from the output layer or Inception5 may loss discriminative power to either visual details or semantic cues. To preserve both semantics and visual details, we propose to generate Semantic Feature Maps (SFMs) from Inception5 and generate visual words from SFMs.

\begin{figure}[t]
\begin{minipage}[b]{1.0\linewidth}
  \centering
  \centerline{\epsfig{figure=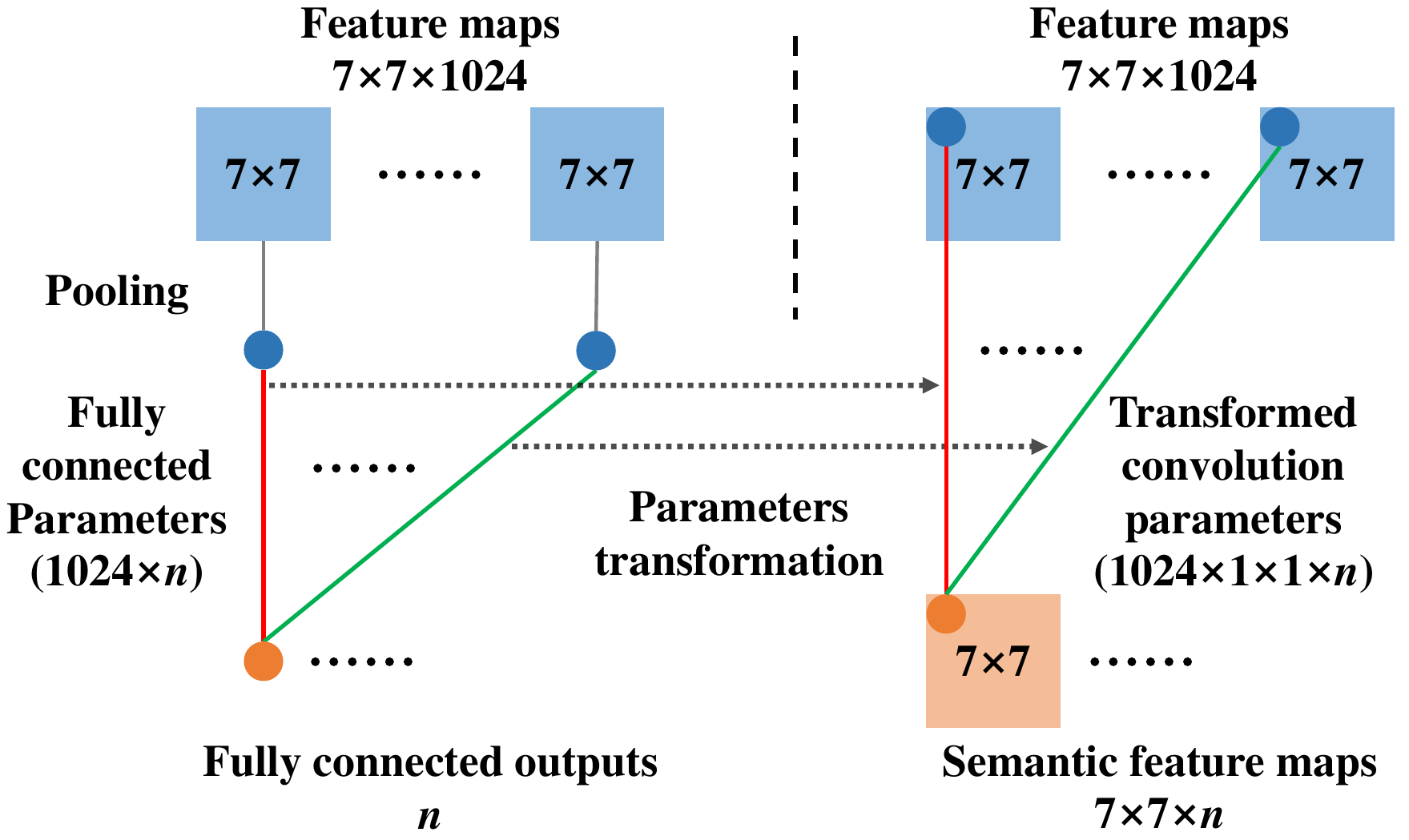,width=6cm}}
\end{minipage}
\setlength{\abovecaptionskip}{-5pt}
\caption{Illustration of transforming parameters of FC layer into a convolutional layer to generate SFMs. Lines in same color indicate the same parameters.}
\label{fig:fc2conv}
\end{figure}

SFMs is generated by transforming the parameters in FC layers into a convolutional layer. This transformation is illustrated in Fig. \ref{fig:fc2conv}. The size of parameters in the FC layer is $ 1024 \times n $, where 2014 is the feature dimensionality after pooling and $n$ is the number of training categories. Those parameters can be reshaped into $ n $ convolutional kernels of size $ 1024 \times 1 \times 1 $. In other words, we transform parameters corresponding to each output in FC layer of size $ 1024 \times 1 $ into a convolutional kernel of size $ 1 \times 1 \times 1024$. Therefore, $n$-channels of convolutional kernel can be generated. Accordingly, $n$ SFMs can be generated after Inception5.

In FC layers, each output is a classification score for an object category. Compared with the output of FC layer, SFMs also contain such classification cues. For example, average pooling the activation on each SFM gets the classification score for the corresponding category. Moreover, SFMs preserve certain visual cues because they are produced from Inception5 without pooling.

\begin{figure}[t]
\begin{minipage}[b]{1.0\linewidth}
  \centering
  \centerline{\epsfig{figure=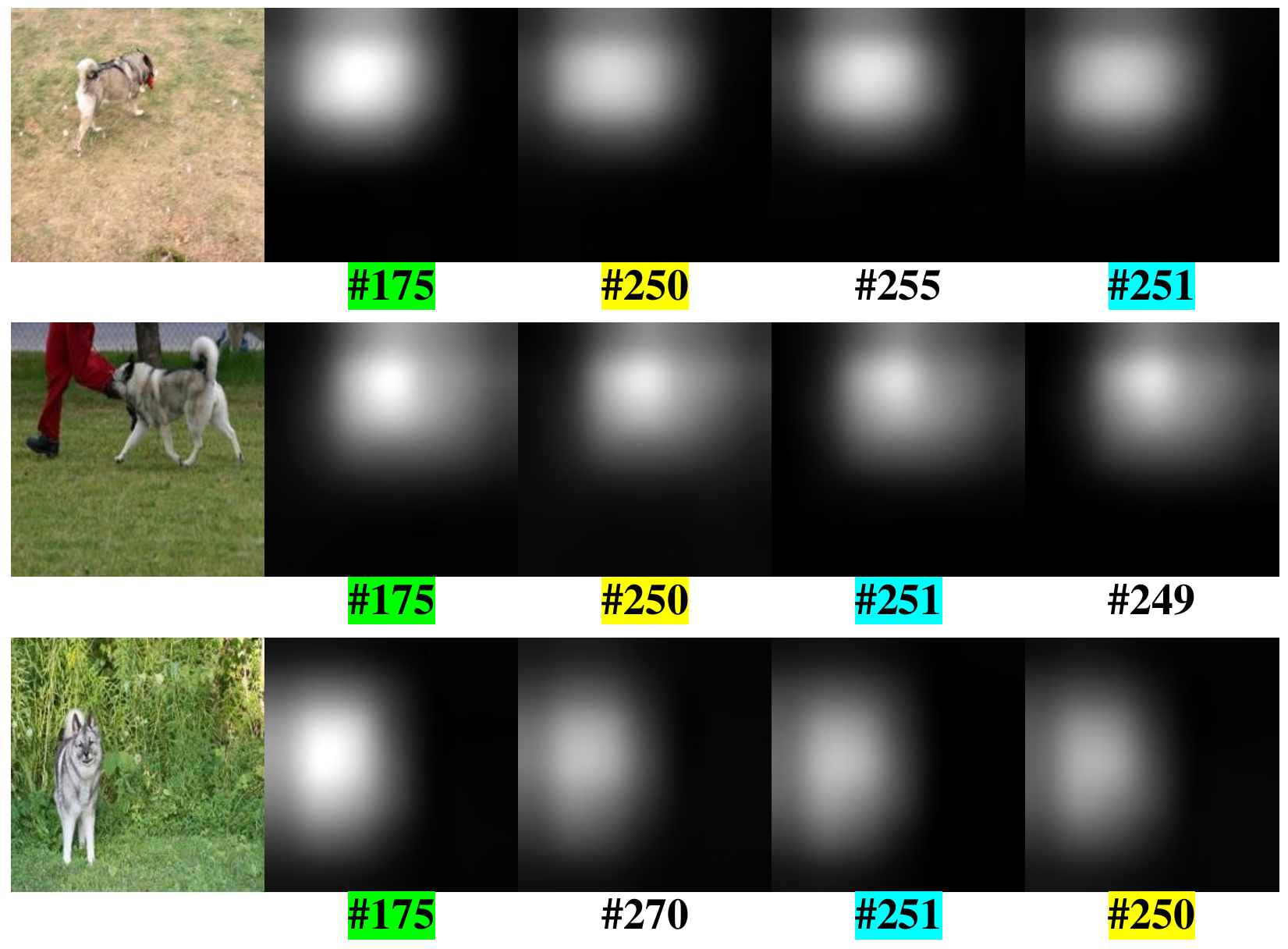,width=6cm}}
\end{minipage}
\setlength{\abovecaptionskip}{-5pt}
\caption{Visualization of some SFMs. Input images are in the first column. The rest are SFMs with top-4  largest response values. The number under each SFM denotes its unique ID in all SFMs. The same IDs are highlighted with the same color.}
\label{fig:featuremap}
\end{figure}

We illustrate examples of SFMs in Fig. \ref{fig:featuremap}. Three images with the same label ``elkhound" in \emph{ImageNet}~\cite{imagenet} and their SFMs with the top-4 largest response values are illustrated. It can be observed that, the illustrated SFMs show 75\% overlap among the three images. SFM \#175 constantly shows the strongest activation. This means the activation values of SFMs represent the semantic and category cues. Moreover, the location and size of object are presented by SFMs.

\subsection{Bag-of-Words Layer}
\label{sec:bowlayer}
\begin{table*}[t]
\begin{center}
\caption{Retrieval efficiency and accuracy on \emph{CIFAR-100}~\cite{cifar} testing set with different thresholds.}
\begin{tabular}{*{12}{c}c}
\hline
Threshold & $0$ & $0.05$ & $0.06$ & $0.07$ & $0.08$ & $0.09$ & $0.10$ & $0.11$ & $0.13$ & $0.13$ & $0.14$ & $0.15$\\
\hline
mAP  & $0.697$ & $0.686$ & $0.689$ & $0.693$ & $0.697$ & $0.700$ & $0.703$ & $\bm{0.704}$ & $0.703$ & $0.700$ & $0.693$ & $0.684$  \\
ANV  & $409.0$ & $50.4$ & $36.7$ & $28.4$ & $23.0$ & $19.0$ & $16.8$ & $15.0$ & $13.5$ & $12.3$ & $11.4$ & $10.6$ \\
ANI  & $4090$ & $500$ & $370$ & $280$ & $230$ & $190$ & $170$ & $150$ & $140$ & $120$ & $110$ & $110$ \\
AOP  & $1,672,810$ & $25,200$ & $13,579$ & $7,952$ & $5,290$ & $3,610$ & $2,856$ & $2,250$ & $1,890$ & $1,476$ & $1,254$ & $1,166$ \\
\hline
\end{tabular}
\label{tab:sparsity}
\end{center}
\end{table*}
\setlength{\textfloatsep}{1pt}

Because different SFMs correspond to different object categories, they are potential to identify and separate the objects in images. Those characteristics make SFMs more suitable to generate visual words that conveys both semantic and visual cues. To preserve the spacial and semantic cues in SFMs, we introduce Bag-of-Words Layer (BoWL) to generate sparse visual words directly from each individual SFM.

Specifically, a local FC layer with ReLU is used to generate $m$ visual words from each individual SFM. This strategy finally generates $m \times n$ visual words. Each local FC layer is trained independently. Compared with traditional FC layer, local FC layer better preserves semantic and visual cues in each SFM, and introduces less parameters to learn. For example, BoWL needs $49 \times m \times n$ parameters, while a FC layer following a pooling layer needs $(49 \times n) \times (m \times n)$ parameters. It should be noted that, we discard SFMs with negative average active values during visual words generation. This reduces the number of nonzero visual words and improves the efficiency for indexing and retrieval.

The generated visual words are L2-normalized for inverted file indexing and retrieval. Our experiments show that, there commonly exist many visual words with small response values, \emph{e.g.}, $1e$-$3$. During online retrieval, those visual words won't contribute much to the similarity computation. Moreover, they are harmful to the sparsity of the BoWs model and would make more images embedded in inverted lists, resulting in more memory overhead. We find that discarding visual words, whose response values are smaller than a threshold, dramatically improves the retrieval efficiency without degrading the accuracy. This procedure is formulated as follows:
\begin{eqnarray}
& thr(x,\beta)=\left\{
\begin{aligned}
  x &,& x>\beta\\
  0 &,& otherwise &
\end{aligned}
\right.
\label{eqn:threshold}
\end{eqnarray}
where $\beta$ denotes the threshold.

We evaluate this procedure on the testing set of \emph{CIFAR-100}~\cite{cifar} with different thresholds. We measure the retrieval performance by mean Average Precision (mAP). The efficiency is measured by Average Number of Operation (ANO) per query image. Using inverted file index, ANO can be approximately computed as the product of Average Number of nonzero Visual words per image (ANV) and Average Number of Images in each inverted list (ANI), \emph{i.e.}, ANO$=$ANV$\times$ANI. Therefore, large mAP implies high discriminative power, and small ANO implies high efficiency for indexing and retrieval. The results are shown in Tab. \ref{tab:sparsity}. It is clear that, retrieval efficiency is significantly improved by filtering visual words with small response values. Meanwhile, retrieval accuracy is improved by removing noisy visual words.

In the aforementioned procedure, the threshold is hard to decide for different testing sets. To determine the threshold automatically, we design a sparsity loss function based on KLD as following:
\begin{eqnarray}
\ell_{spa}(\beta) = \hat{\rho}\log\frac{\hat{\rho}}{\rho} + (1-\hat{\rho})\log\frac{(1-\hat{\rho})}{1-\rho},
\label{eqn:sparsityloss}
\end{eqnarray}
where $\hat{\rho}$ denotes the desired ratio between the number of nonzero visual words and the total number of visual words. $\rho$ is the ratio computed on training set of $N$ images, \emph{i.e.},
\begin{eqnarray}
\rho = \frac{1}{N \times m \times n}\sum_{i=1}^{N}\sum_{j=i}^{m \times n}sign(v_i(j)-\beta).
\end{eqnarray}
$sign(\cdot)$ is sign function defined as follows:
\begin{eqnarray}
& sign(x)=\left\{
\begin{aligned}
  1 &,& x>0\\
  0 &,& otherwise &
\end{aligned}
\right.
\label{sparsity}
\end{eqnarray}
With this object function, the model is trained to learn the threshold $\beta$ to ensure a ratio of $\hat{\rho}$ visual words are nonzero. We thus use this sparsity loss to control the sparsity of the generated visual words.

\subsection{Model Training}
The overall network is trained by SGD with object function as following,
\begin{eqnarray}
L(\theta,\beta)=\ell_{cls}+\lambda_1\ell_{tri}+\lambda_2\ell_{spa},
\end{eqnarray}
where $\theta$ denotes parameters in convolutional layers, $\beta$ denotes the threshold in BoWL, $\ell_{cla}$, $\ell_{tri}$ and $\ell_{spa}$ denote the loss of classification, triplet similarity and sparsity, respectively. Since only using the triplet loss takes a long time to converge, we further introduce the classification loss to ensure fast convergence. The triplet similarity loss ensures the discriminative ability of the learned features in similarity computation. The sparsity loss ensures retrieval efficiency.

We design the triplet similarity loss as:
\begin{eqnarray}
\ell_{tri}(v_a, v_p, v_n)=max\{0, sim_{v_a}^{v_n} -
sim_{v_a}^{v_p} + \alpha \},
\label{eqn:simloss}
\end{eqnarray}
where $\alpha$ is the margin parameter, $v_a$, $v_p$ and $v_n$ are the vectors of L2-normalized visual words of anchor image, similar image, and dissimilar image, respectively. $sim_{v_1}^{v_2}$ is the cosine distance between two vectors, \emph{i.e.},
$sim_{v_1}^{v_2} = v_1^T * v_2$.
When $\ell_{tri}(v_a, v_p, v_n) \neq 0$, the gradient with respect to each vector can be computed as:
\begin{eqnarray}
\frac{\partial \ell_{tri}(v_a, v_p, v_n)}{v_a}&=&v_n - v_p \\
\frac{\partial \ell_{tri}(v_a, v_p, v_n)}{v_p}&=&-v_a \\
\frac{\partial \ell_{tri}(v_a, v_p, v_n)}{v_n}&=&v_a
\end{eqnarray}

Different from other works that use Euclidean distance to compute the triplet similarity, we choose Cosine distance to make similar images share more visual words and vice versa. This is mainly because we also use Cosine distance to compute image similarity during retrieval based on inverted indexes.

The sparsity loss $\ell_{spa}$ is formulated in Eq. \ref{eqn:sparsityloss}. Since the $sign(\cdot)$ function is non-differential, we define the gradient of it as
\begin{eqnarray}
 \frac{\partial sign(v_i(j)-\beta)}{\partial \beta}　& = & -sign(v_i(j)-\beta) \nonumber \\
 & = & \left\{
\begin{aligned}
  -1 &,& v_i(j)-\beta >0\\
  0  &,& otherwise &
\end{aligned}
\right.
\label{signgrad}
\end{eqnarray}
The gradient of $\ell_{spa}(\beta)$ can be computed as
\begin{eqnarray}
 \frac{\partial \ell_{spa}(\beta)}{\partial \beta} & = & \frac{\partial \ell_{spa}(\beta)}{\partial \rho} \cdot \frac{\partial \rho}{\partial \beta} \nonumber \\
 & = & \frac{\hat{\rho}-\rho}{1-\rho}
\label{sparsegrad}
\end{eqnarray}
Therefore, $\beta$ can be leaned by gradient descent method.

\subsection{Generalization Ability Improvement}

\begin{figure}[t]
\begin{minipage}[b]{1.0\linewidth}
  \centering
  \centerline{\epsfig{figure=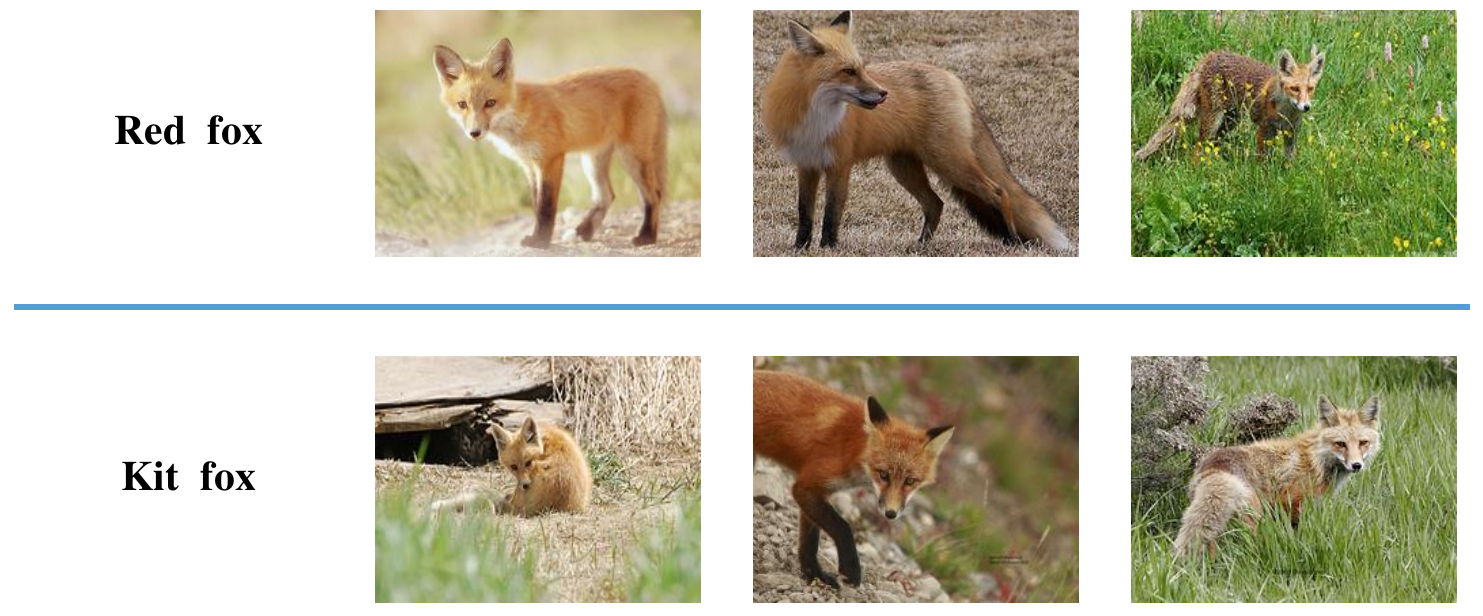,width=6cm}}
\end{minipage}
\setlength{\abovecaptionskip}{-5pt}
\caption{Illustration of two categories in \emph{ImageNet}~\cite{imagenet}, that are visually and semantically similar.}
\label{fig:imagenet}
\end{figure}

Most of conventional retrieval models based on DCNN need to be fine-tuned on the domain dataset~\cite{wan}. However, fine-tuning is commonly unavailable in real image retrieval applications. Then \emph{ImageNet}~\cite{imagenet} could be a reasonable option for training as it contains large-scale labeled images. However, \emph{ImageNet} contains some fine-grained categories and some categories are both visually and semantically similar as shown in Fig. \ref{fig:imagenet}.

In our method, different categories correspond to different SFMs, which hence generate different visual words. It's not reasonable to regard similar categories to generate unrelated visual words, when using \emph{ImageNet} as the training set. For example, images of ``red fox" should be allowed to share more visual words with images of ``kit fox" than with images of ``jeep". Therefore, original labels in \emph{ImageNet}~\cite{imagenet} are not optimal for training E$^2$BoWs and may mislead the model for retrieval tasks.

To tackle the above issue, we change the parameter $\alpha$ in triplet loss function according to the similarity of two categories, \emph{i.e.}, set a small value of $\alpha$ for images of similar categories and use a large value for images of dissimilar categories. Specifically, we first compute the similarity between two categories based on the tree struct\footnote{ImageNet Tree View. http://image-net.org/explore} of \emph{ImageNet}~\cite{imagenet}. Given $H$ denotes the height of the tree and $h_{c_1}^{c_2}$ denotes the height of the common parent nodes of two different categories $c_1$ and $c_2$, the similarity $S(c_1,c_2)$ between $c_1$ and $c_2$ is defined as: $S(c_1,c_2)=\frac{h}{H}$.
Then we modify parameter $\alpha$ as:
\begin{eqnarray}
\alpha'=\frac{\alpha}{(1+S(c_1,c_2))^2}
\label{modify-alpha}
\end{eqnarray}
The above strategy allows images from similar categories to share more common visual words, thus makes $ImageNet$ a more reasonable training set. It is thus potential to improve the generalization ability of the learned E$^2$BoWs on other unseen datasets.

\section{Experiments}
\label{sec:exper}

\subsection{Datasets and Implementation Details}

We first evaluate our model in tiny image retrieval task on \emph{CIFAR-10}~\cite{cifar} and \emph{CIFAR-100}~\cite{cifar}. Then, our model is evaluated in image retrieval task on \emph{MIRFLICKR-25K}~\cite{flickr}. Finally, we compare the generalization ability between the proposed E$^2$BoWs and deep features extracted from GoogLeNet~\cite{googlenet} without/with BN~\cite{bn} by firt training the model on \emph{ImageNet}~\cite{imagenet} and then testing the model on \emph{NUS-WIDE}~\cite{nus}. Details of those test sets are given as follows:
\begin{itemize}
  \item\emph{CIFAR-10}~\cite{cifar} contains tiny images belonging to 10 classes. Each class contains 5,000 training images and 1,000 testing images.
  \item\emph{CIFAR-100}~\cite{cifar} contains 100 classes of tiny images. Each class contains 500 training images and 100 testing images. Retrieval task on it is more challenging than the one on \emph{CIFAR-10}~\cite{cifar}.
  \item\emph{MIRFLICKR-25K}\cite{flickr} consists of 25,000 images with 38 concepts.
  \item\emph{ImageNet}~\cite{imagenet} contains 1,000 categories and around 1,200 images per category.
  \item\emph{NUS-WIDE}~\cite{nus} consists of around 270K images and 81 concepts.
\end{itemize}

Each SFM corresponds to a category on the training set. Therefore, the number of SFMs equals to the number of training categories. On \emph{CIFAR-10}, \emph{CIFAR-100}, and \emph{MIRFLICKR-25K}, 10 visual words are generated from each SFM. This results in 100, 1,000 and 380 visual words, respectively. For \emph{ImageNet}~\cite{imagenet}, we generate 25 visual words on each SFM and get totally 25,000 visual words. Margin parameter in similarity loss is set to 0.2 on all datasets.

mAP (mean Average Precision) is used to evaluate the retrieval performance on \emph{CIFAR-10}, \emph{CIFAR-100}, and \emph{NUS-WIDE}. For \emph{MIRFLICKR-25K}~\cite{flickr}, we use NDCG@100 as the evaluation metric to consider different levels of relevance. In Tab. \ref{tab:results}, \ref{tab:results2}, and \ref{tab:generalization}, the tag ``-B" denotes that feature is binarized by using $sign(\cdot)$ function to accelerate the retrieval.

\subsection{Performance on CIFAR}

On \emph{CIFAR-10} and \emph{CIFAR-100}, we use the training sets for model fine-tuning and use the test sets for retrieval, respectively. The sparsity loss parameter $\hat{\rho}$ is set as 0.08 and 0.01 on \emph{CIFAR-10} and \emph{CIFAR-100}, respectively depending on the number of categories. We compare the retrieval performance between E$^2$BoWs and existing methods including ITQ~\cite{itq}, ITQ-CCA~\cite{itq}, KSH~\cite{ksh}, SH~\cite{sh}, MLH~\cite{mlh}, BRE~\cite{bre}, CNNH~\cite{cnnh}, CNNH+~\cite{cnnh}, DNNH~\cite{dnnh}, DSH~\cite{dsh}, and BHC~\cite{linkevin}.

The performance comparison is summarized in Tab. \ref{tab:results}, which shows the best performance of each method with 48-bit codes. The compared methods do not report their performance on the \emph{CIFAR-100}. Among those methods, BHC~\cite{linkevin} shows the best performance on \emph{CIFAR-10}. Therefore, we implement BHC~\cite{linkevin} and report its performance on \emph{CIFAR-100} for comparison. In Tab. \ref{tab:results}, ``*'' denotes our implementation. It can be observed from Tab. \ref{tab:results} that, methods based on DCNN perform better than conventional retrieval methods using hand-crafted features. Among DCNN based methods, our model yields the highest mAP on the two datasets. It is also clear that, our work also show substantial advantage on the more challenging \emph{CIFAR-100}~\cite{cifar} dataset.

\begin{table}[t]
\begin{center}
\caption{Comparison of mAP (\%) among different methods on \emph{CIFAR-10}~\cite{cifar} and \emph{CIFAR-100}~\cite{cifar}.}
\begin{tabular}{c*{2}{c}}
\hline
Method  & CIFAR-10 &  CIFAR-100 \\
\hline
ITQ~\cite{itq}     & $0.175$  & ---\\
ITQ-CCA~\cite{itq} & $0.295$  & --- \\
KSH~\cite{ksh}     & $0.315$ & --- \\
SH~\cite{sh}       & $0.132$ & --- \\
MLH~\cite{mlh}     & $0.211$ & --- \\
BRE~\cite{bre}     & $0.196$ & --- \\
\hline
CNNH~\cite{cnnh}    & $0.522$ & --- \\
CNNH+~\cite{cnnh}   & $0.532$  & ---\\
DNNH~\cite{dnnh}    & $0.581$ & --- \\
DSH~\cite{dsh}      & $0.676$  & --- \\
BHC~\cite{linkevin}   & $0.897$ & $0.650^*$ \\
\hline
E$^2$BoWs    & $\bm{0.909}$ & $\bm{0.689}$  \\
E$^2$BoWs-B  & $0.908$ & $0.624$  \\
\hline
\end{tabular}
\label{tab:results}
\end{center}
\end{table}

\subsection{Performance on MIRFLICKR-25K}

\begin{table}[t]
\begin{center}
\caption{Comparison of NDCG@100 among different methods on \emph{MIRFLICKR-25K}~\cite{flickr}.}
\begin{tabular}{ccc|cc}
\hline
ITQ-CCA~\cite{itq} &  KSH~\cite{ksh} & BHC~\cite{linkevin} & E$^2$BoWs & E$^2$BoWs-b \\
\hline
 $0.402$  & $0.350$ & $0.510^*$ & $0.492$ & $\bm{0.526}$ \\
\hline
\end{tabular}
\label{tab:results2}
\end{center}
\end{table}
\setlength{\textfloatsep}{3pt}

\begin{figure*}[t]
\begin{minipage}{1\linewidth}
  \centering
  \centerline{\epsfig{figure=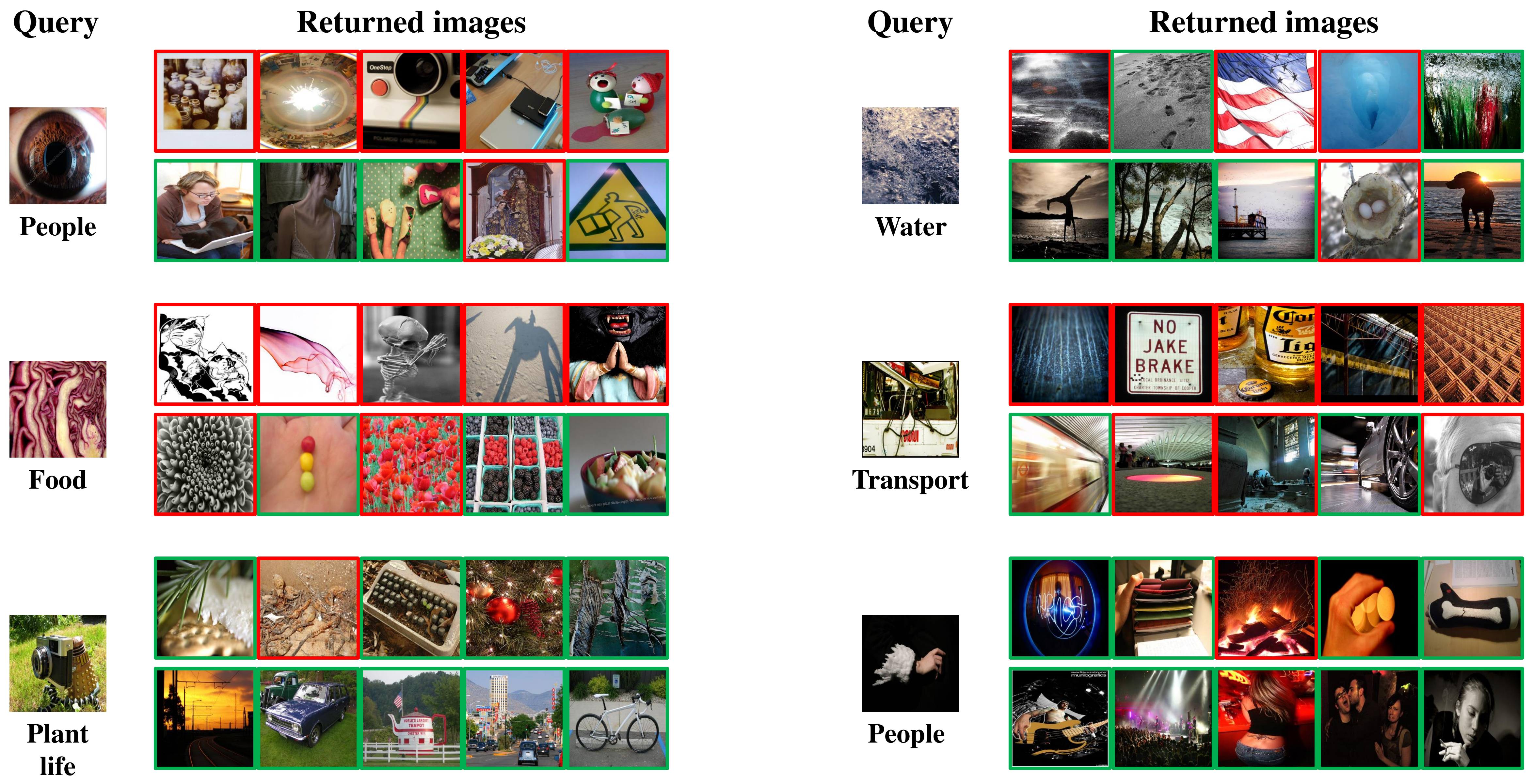,width=13cm}}
\end{minipage}
\captionsetup{belowskip=-3pt}
\caption{Examples of retrieval results of BHC~\cite{linkevin} and proposed E$^2$BoWs-B on \emph{MIRFLICKR-25K}~\cite{flickr}. In each example, the query image is placed on the left with the ground truth label under it. The first row shows the top 5 images returned by BHC~\cite{linkevin}, the second row shows the result of E$^2$BoWs-B. Relevant/irrelevant images are annotated by green/red boxes, respectively.}
\label{fig:return}
\end{figure*}
\setlength{\floatsep}{10.0pt}

On \emph{MIRFLICKR-25K}~\cite{flickr}, we follow the experimental setting of ~\cite{zhao2015}, where 2,000 images are randomly selected as query images and the rest are used for training. We set sparsity loss parameter $\hat{\rho}$ to 0.11. We also implement BHC~\cite{linkevin} for comparison because it shows the best performance among the compared works on \emph{CIFAR-10}~\cite{cifar}.

Performance comparison is shown in Tab. \ref{tab:results2}. It can be observed that, DCNN based methods also perform better than the conventional methods. This implies the powerful feature learning ability of deep models. It is also clear that, binarized E$^2$Bows achieves the best performance. Examples of image retrieval results of BHC~\cite{linkevin} and E$^2$BoWs-B are shown in Fig. \ref{fig:return}. As shown in Fig. \ref{fig:return}, E$^2$BoWs-B is more discriminative to semantic cues. For example, E$^2$BoWs effectively identifies the semantic of ``people" from an human eye image, and gets better retrieval results than BHC~\cite{linkevin}.

\subsection{Evaluation on Generalization Ability}

\begin{table*}[t]
\begin{center}
\caption{Comparison of mAP (\%) between GoogLeNet feature and E$^2$BoWs on \emph{NUS-WIDE}~\cite{nus}. The compared features are trained on an independent training set.}
\begin{tabular}{c|ccccc|ccccc}
\hline
Feature  & GN$_{1024}$ & GN$_{1000}$ & GN$_{1024}^{BN}$ & GN$_{1000}^{BN}$ & E$^2$BoWs & GN$_{1024}$-B & GN$_{1000}$-B & GN$_{1024}^{BN}$-B & GN$_{1000}^{BN}$-B & E$^2$BoWs-B \\
\hline
mAP      & $ 0.552 $ & $0.594$ & $ 0.551 $ & $0.591$ & $\bm{0.599}$ & $0.388$ & $0.549$ & $0.326$ & $0.543$ &  $\bm{0.563}$\\
\hline
\end{tabular}
\label{tab:generalization}
\end{center}
\end{table*}
\setlength{\floatsep}{10.0pt}

To validate the generalization ability of the proposed E$^2$BoWs feature, we first train E$^2$BoWs on \emph{ImageNet}~\cite{imagenet}, then test it on \emph{NUS-WIDE}~\cite{nus}. When training on \emph{ImageNet}~\cite{imagenet}, the sparsity loss parameter is relaxed to 0.14 and 25 visual words are generated from each SFM. The retrieval on \emph{NUS-WIDE}~\cite{nus} uses the same experimental setting in~\cite{dsh, cnnh}, \emph{i.e.}, use the images associated with the 21 most frequent concepts and the testing set in~\cite{dsh}, which consists of 10,000 images. As one image may be associated with many concepts, we follow~\cite{dsh} and consider two images are similar if they share at least one concept. We compare our model with features generated directly from GoogLeNet~\cite{googlenet} with and without BN~\cite{bn}, \emph{i.e.},
\begin{itemize}
  \item GN$_{1024}$/GN$_{1024}^{BN}$: 1024-d feature extracted from the pool5 layer in GoogLeNet~\cite{googlenet} without/with BN~\cite{bn}.
  \item GN$_{1000}$/GN$_{1000}^{BN}$: 1000-d feature extracted from the output layer in GoogLeNet~\cite{googlenet} without/with BN~\cite{bn}.
\end{itemize}

The comparison between E$^2$BoWs and GoogLeNet features is summarized in Tab. \ref{tab:generalization}. It could be observed that our model constantly shows better retrieval accuracy. Note that, the above experiments use independent training and testing sets. Therefore, we can conclude that E$^2$BoWs shows better generalization ability than GoogLeNet features.

\subsection{Discussions}

During training, we encourage E$^2$BoWs to be sparse to ensure its high efficiency in inverted file indexing and retrieval. On \emph{CIFAR-10}, \emph{CIFAR-100}, and \emph{MIRFLICKR-25K}, we analyze the retrieval complexity of our E$^2$BoWs model and compare it with the one of 48-bit binary code generated by BHC~\cite{linkevin}.

As shown in Tab. \ref{tab:efficiency}, E$^2$BoWs is sparse. For instance, the average number of visual words in each image on \emph{MIRFLICKR-25K} is about 44, which is significantly smaller than the total visual word size 380. It is also clear that, with inverted file index, retrieval based on E$^2$BoWs can be efficiently finished with less operations than the linear search with binary code. From the above experiments, we can conclude E$^2$BoWs shows advantages in the aspects of both accuracy and efficiency, compared with BHC~\cite{linkevin}.
\begin{table}[t]
\begin{center}
\caption{Retrieval efficiency of different methods on \emph{CIFAR-10}~\cite{cifar}, \emph{CIFAR-100}~\cite{cifar} and \emph{MIRFLICKR-25K}~\cite{flickr}.}
\begin{tabular}{c*{4}{c}}
\hline
Method  & & CIFAR-10 &  CIFAR-100 & \emph{MIRFLICKR-25K} \\
\hline
BHC~\cite{linkevin}& ANO & $480,000$ & $480,000$ & $406,944$ \\
\hline
\multirow{3}{*}{E$^2$BoWs} & ANV & $8.64$ & $10.6$ &$43.6$ \\
 & ANI & $960$ & $110$ & $975$ \\
 & ANO  & $8,294$ & $1,166$ & $42,510$ \\
\hline
\end{tabular}
\label{tab:efficiency}
\end{center}
\end{table}
\setlength{\textfloatsep}{3pt}

\section{Conclusions}
\label{sec:concl}
This paper presents E$^2$BoWs for large-scale CBIR based on DCNN. E$^2$BoWs first transforms FC layer in GoogLeNet~\cite{googlenet} into convolutional layer to generate semantic feature maps. Visual words are then generated from these feature maps by the proposed Bag-of-Words layer to preserve both the semantic and visual cues. A threshold layer is hence introduced to ensure the sparsity of generated visual words. We also introduce a novel learning algorithm to reinforce the sparsity of the generated E$^2$BoWs model, which further ensures the time and memory efficiency. Experiments on four benchmark datasets demonstrate that our model shows substantial advantages in the aspects of discriminative power, efficiency, and generalization ability.

\section*{Acknowledgements}
This work is supported by National Science Foundation of China under Grant No. 61572050, 91538111, 61620106009, 61429201, and the National 1000 Youth Talents Plan, in part to Dr. Qi Tian by ARO grant W911NF-15-1-0290 and Faculty Research Gift Awards by NEC Laboratories of America and Blippar.

\bibliographystyle{IEEEtran}
\bibliography{IEEEabrv,tran}
\end{document}